\documentclass{article}
\usepackage{arxiv}
\usepackage[utf8]{inputenc}
\usepackage[T1]{fontenc}
\usepackage{hyperref}
\usepackage{url}
\usepackage{booktabs}
\usepackage{amsfonts,amssymb,amsmath}
\usepackage{nicefrac}
\usepackage{microtype}
\usepackage{graphicx}
\usepackage{float}
\usepackage{enumitem}
\usepackage{cite}
\graphicspath{ {./images/} }

\title{An Optimized Machine Learning Classifier for Detecting Fake Reviews Using Extracted Features}

\author{
  Anshuman \\
  Indian Institute of Information Technology Vadodara\\
  \texttt{202351009@iiitvadodara.ac.in} \\
  \And
  Ayush Chaurasia \\
  Indian Institute of Information Technology Vadodara\\
  \texttt{202351015@iiitvadodara.ac.in} \\
  \And
  Bogar Prathmesh Balaji \\
  Indian Institute of Information Technology Vadodara\\
  \texttt{202351019@iiitvadodara.ac.in} \\
  \And
  Shabbir Khuzem Anees \\
  Indian Institute of Information Technology Vadodara\\
  \texttt{202352331@iiitvadodara.ac.in} \\
}

\begin{document}
\maketitle

\begin{abstract}
It is well known that fraudulent reviews cast doubt on the legitimacy and dependability of online purchases. The most recent development that leads customers towards darkness is the appearance of human reviews in computer-generated (CG) ones. In this work, we present an advanced machine-learning-based system that analyses these reviews produced by AI with remarkable precision. Our method integrates advanced text preprocessing, multi-modal feature extraction, Harris Hawks Optimization (HHO) for feature selection, and a stacking ensemble classifier. We implemented this methodology on a public dataset of 40,432 Original (OR) and Computer-Generated (CG) reviews. From an initial set of 13,539 features, HHO selected the most applicable 1,368 features, achieving an 89.9\% dimensionality reduction. Our final stacking model achieved 95.40\% accuracy, 92.81\% precision, 95.01\% recall, and a 93.90\% F1-Score, which demonstrates that the combination of ensemble learning and bio-inspired optimisation is an effective method for machine-generated text recognition. Because large-scale review analytics commonly run on cloud platforms, privacy-preserving techniques such as differential approaches and secure outsourcing are essential to protect user data in these systems.
\end{abstract}


\section{Introduction}

The modern world of e-commerce is heavily dependent on online reviews, which are the key factor driving consumers' buying decisions. However, the credibility of reviews is under threat due to the increasing cases of fraudulent. One major reason for this is the problem of fake reviews written by humans \cite{kumar2020detecting,jindal2008opinion,ott2011finding}, while the latest complex problem is the introduction of \textbf{computer-generated (CG) reviews} \cite{salminen2022creating,gehrmann2019gltr,zellers2019defending,mitchell2023detectgpt}. The latter can generate reviews in large quantities that are frequently so similar to human writing that manual detection is practically impossible \cite{radford2019language,brown2020language}.

The stakes for automated detection systems have increased due to this new threat, necessitating \textbf{"machines to fight machines"} at a much higher level. Even though deep learning models can occasionally yield useful results, they still rely heavily on powerful computers. In order to increase both efficacy, another field of study has pointed on \textbf{feature-based machine learning} and used \textbf{bio-inspired optimisation algorithms} such as \textbf{Harris Hawks Optimisation (HHO)} \cite{mirjalili2019genetic}. \textbf{Fadhel et al.} \cite{fadhel2022optimized} shows the effectiveness of HHO in feature selection by implementing it flawlessly in the field of fake review detection.

Additionally, the storage and distributed processing of large datasets, \textbf{cloud computing} has been instrumental in the implementation of scalable review-analysis systems \cite{armbrust2010cloud,marinescu2017cloud}.  But as these systems grow, concerns about secure computation and \textbf{data privacy} become crucial, particularly when working with AI-generated text and user-generated reviews \cite{zhang2010privacy,shokri2015privacy,dwork2008differential}. Recent work specifically addresses differential privacy and access-policy mechanisms for privacy-preserving cloud-based ML and data-sharing approaches that are directly applicable to large-scale review analytics \cite{dp1,dp4,dp3}. Secure data sharing protocols and attribute-based encryption are complementary tools for protecting user data in outsourced/cloud settings \cite{dp6,dp7}.

To address the unique problem of CG review detection, we suggest a new system that makes use of the \textbf{HHO methodology} \cite{fadhel2022optimized}. Our main contributions are:
\begin{enumerate}
    \item The creation of a large, \textbf{multi-modal feature set} (13,539 features) \cite{wang2012detecting,pham2016feature}.
    \item The adoption of \textbf{HHO} for efficient dimensionality reduction \cite{fadhel2022optimized}.
    \item The use of a \textbf{stacking ensemble classifier} \cite{mirjalili2019genetic,van2007super}.
    \item Examining privacy-preserving deployment alternatives for cloud-based review analytics pipelines, including fog-assisted storage, access controls, attribute-based encryption, and differentiated methods \cite{dp2,dp5,dp6,dp9}.
\end{enumerate}

\section{Related Work}

Our research is mainly dependent on two main areas: firstly, the definition of the computer-generated review issue and secondly, the utilization of \textbf{bio inspired optimization} for feature selection.

\textbf{Salminen et al.} \cite{salminen2022creating} set apart the mimicked human-written fake reviews and the AI-generated text as a new danger. Their research consists the use of language models (like \textbf{GPT-2} \cite{radford2019language}) to generate a huge collection of artificial reviews. The most important outcome of their research was that human reviewers had almost random luck in identifying the CG reviews (\textbf{55\% accuracy}), whereas a machine classifier could have the same level of detection as human annotators. This led to the establishment of the dataset as an important benchmark and the verification of the necessity of advanced automated techniques in this field.

As well as, \textbf{Kumar et al.} \cite{kumar2020detecting} investigated linguistic and behavioral signals to find fake online content and stated that combining \textbf{text-based} and \textbf{metadata-based} signals improved the dependence of detection. Likewise, \textbf{Zhang et al.} \cite{zhang2010privacy} proposed deep contextual embedding models to differentiate between AI-generated and human-written text, focusing the importance of \textbf{hybrid methods} that join traditional feature engineering and machine learning approaches \cite{devlin2019bert,liu2019roberta,nguyen2020adversarial}.

The observation of fake text often requires handling \textbf{high-dimensional feature spaces}, which can be computationally costly and tend to overfit. \textbf{Fadhel et al.} \cite{fadhel2022optimized} directed this issue by using \textbf{Harris Hawks Optimization (HHO)}, a nature-inspired algorithm, for feature selection in the fake review detection. By using HHO, they proceeded by identifying an optimal subset of features that enhanced the performance of classifiers such as \textbf{XGBoost} and \textbf{Extra Trees Classifier (ETC)} on an airline sentiment dataset.

Similar \textbf{bio-inspired techniques} for feature optimization have been disclosed in other studies as well. One typical case is that of \textbf{Mirjalili et al.} \cite{mirjalili2019genetic}, who carried out an evaluation of \textbf{genetic algorithms} versus \textbf{swarm-based optimization} in text classification tasks, demonstrating their efficiency in reducing redundant features.

Beyond fake review detection, related survey and review papers study the broader landscape of fake content, detection techniques, and system design \cite{mewada2022survey}. There is also relevant work on emotion detection from speech and classification which motivates certain feature choices for sentiment and emotion-aware features in textual analysis; for example, literature reviews on emotion detection and specific CNN-based models for emotion classification from speech inform our thinking when designing multimodal and linguistic cues \cite{dp10}. Finally, literature on secure outsourced data models, differential-privacy approaches, and attribute-based encryption provide concrete mechanisms to make cloud-based review analysis privacy-preserving \cite{dp5,dp6,dp7,dp8,dp9,dp11}.

\section{Our Contribution}
Our work synthesizes these research streams by applying HHO-based optimization to the specific problem of detecting computer-generated (CG) reviews. Building on prior findings, we propose a new hybrid pipeline integrating:
\begin{itemize}
    \item A large multi-modal feature set of 13,539 linguistic and statistical features;
    \item Harris Hawks Optimization (HHO) for reducing dimensionality;
    \item A ensemble stacking of ETC, RF, XGBoost, and SVM models.
\end{itemize}
This combination achieves 95.40\% accuracy on the benchmark dataset, showing that selected feature which are efficient and ensemble learning can outperform deeper and more computationally costly models. Furthermore, we address deployment-time privacy safeguards to safeguard user review data when the pipeline is operating in a cloud or fog environment, such as attribute-based access limits and differential noise addition.

\section{Methodology}

\begin{figure}[htbp]
\centerline{\includegraphics[width=\columnwidth]{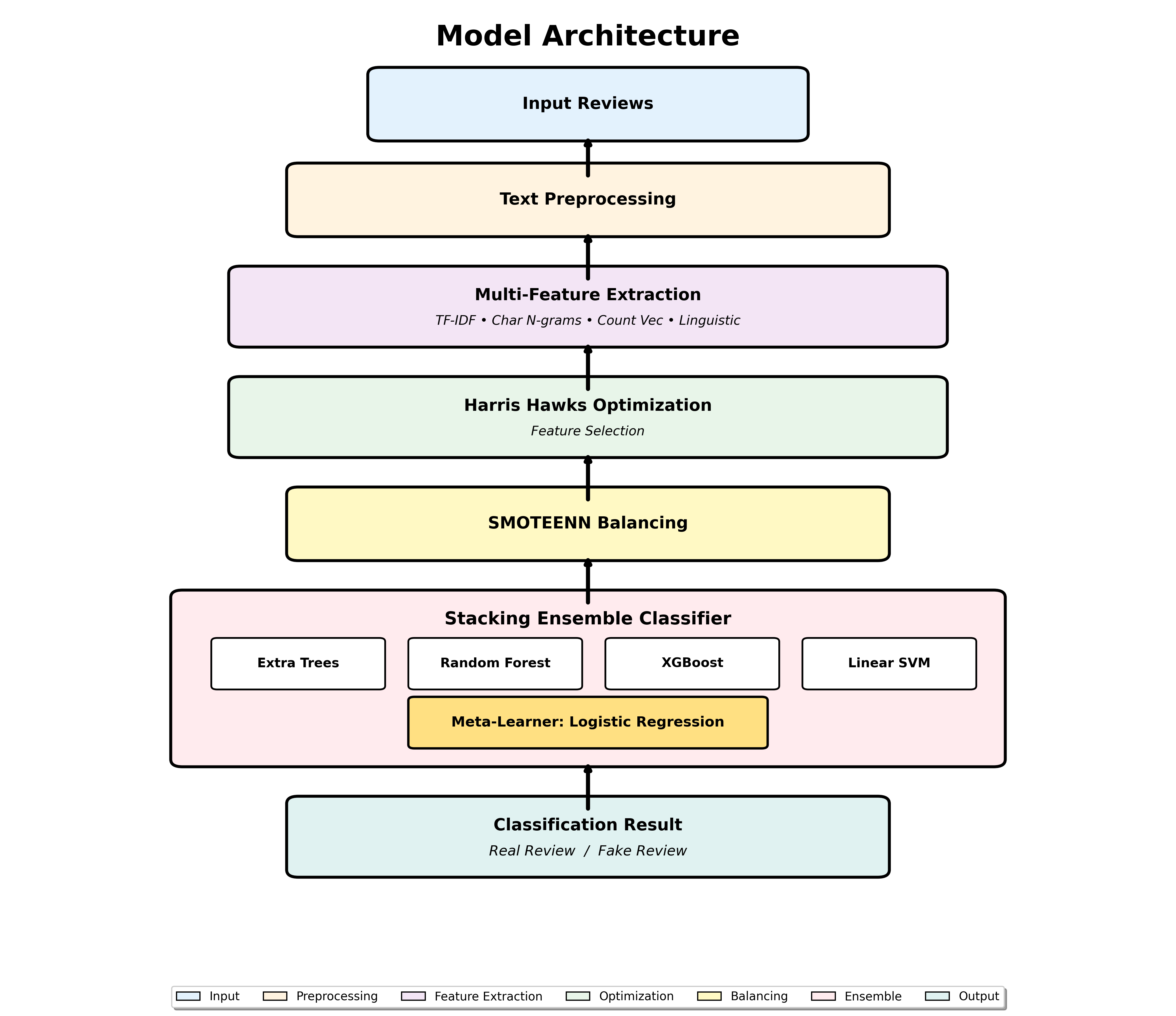}}
\caption{System architecture prediction from input.}
\label{fig:architecture}
\end{figure}

\subsection{Text Preprocessing}
\begin{itemize}
    \item \textbf{Contraction Expansion:} Converting short forms to the standard (such as ``don’t'' to ``do not'').
    \item \textbf{Text Cleaning:} Removing HTML tags, emails, and URLs while smartly saving emotional driven punctuation like “!” \cite{zhang2010privacy,ott2011finding}.
    \item \textbf{Lemmatization:} Bringing words return to their basic forms (eg. ``running'' to ``run'') by the help of WordNet \cite{miller1995wordnet}.
    \item \textbf{Smart Stopword Removal:} Eliminating the similar words while preserving the negations (such as "not" and "never") that are important for sections \cite{kumar2020detecting,li2014towards}.
\end{itemize}

\subsection{Multi-Modal Feature Extraction}

In order to create a complete review "fingerprint," we created four different feature sets totaling \textbf{12,539 dimensions}:

\begin{itemize}
    \item \textbf{TF-IDF Features (10,000):} Highlighting the importance of key words and phrases (1- to 4-grams) \cite{zhang2010privacy,joachims1998text}.
    \item \textbf{Character N-grams (1,500):} Identifying the sub-word patterns (3- to 6-grams) that are majorly of AI text or indicate human presence (e.g., ``!!!'') \cite{gehrmann2019gltr,zhang2010privacy}.
    \item \textbf{Count Vectorizer (2,000):} Representing the most basic word and phrase frequencies (1- and 2-grams).
    \item \textbf{Linguistic Features (43):} Uppercase ratios, word repetition, sentiment polarity (using TextBlob), punctuation density, and other such aspects were counted in features build under human supervision \cite{kumar2020detecting,mukherjee2013spotting}.
\end{itemize}

The result of this multi-modal procedure was a \textbf{13,539 high-dimensional} feature vector.

\subsection{Harris Hawks Optimization}
We used \textbf{Harris Hawks Optimisation (HHO)} \cite{fadhel2022optimized}, a nature-based algorithm that simulates hawks' cooperative hunting behaviour, to handle this high-dimensional data. Using \textbf{30 hawks} for \textbf{50 iterations}, HHO decreased our feature space by \textbf{89.9\%}, from \textbf{13,539} to \textbf{1,368 features}. This is shown in the figure as the removal of noise and the protection of the most discrete signals.

\begin{figure}[htbp]
\centerline{\includegraphics[width=\columnwidth]{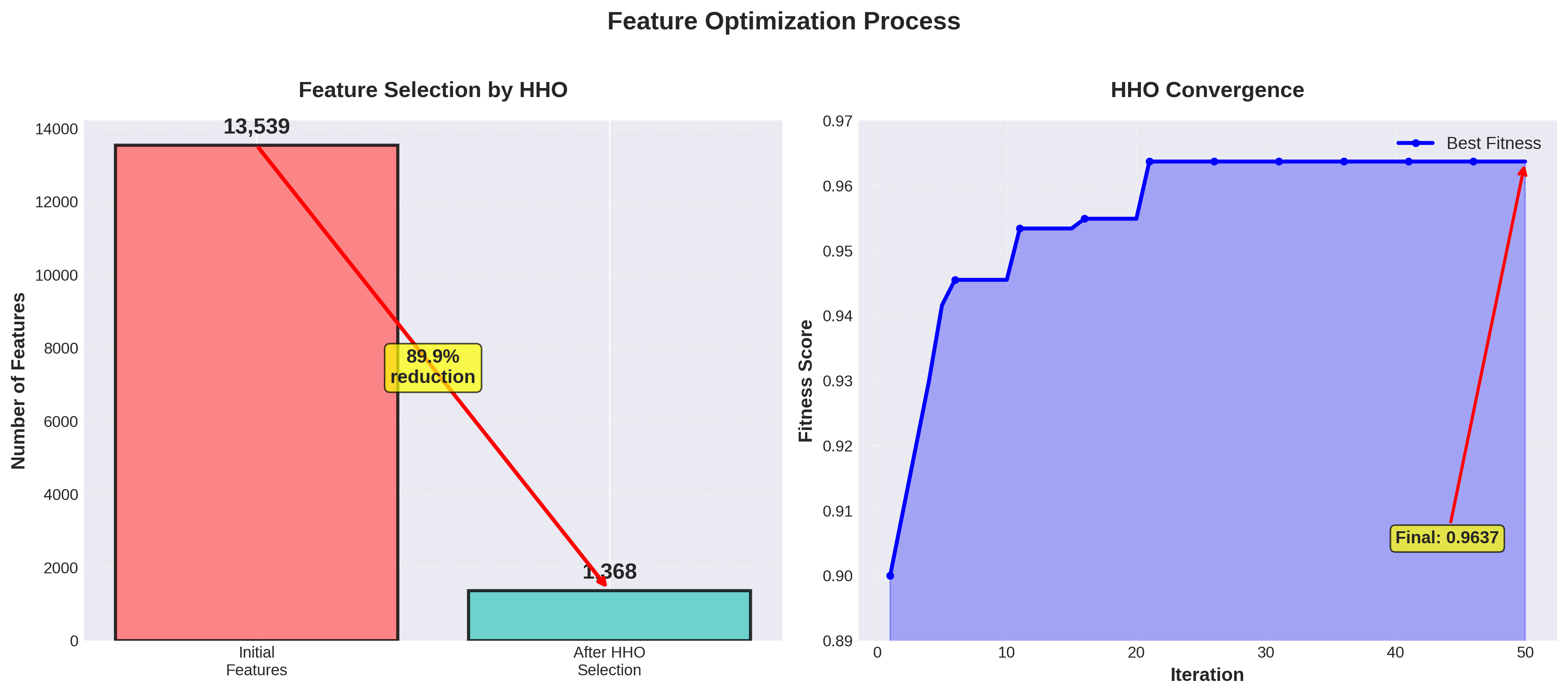}}
\caption{Feature reduction and HHO convergence over 50 iterations.}
\label{fig:feature_selection}
\end{figure}

\subsection{Data Balancing with SMOTEENN}
We used SMOTEENN to clean the training data \cite{chawla2002smote,lemaitre2017imbalanced}. SMOTEENN oversamples the minority class while simultaneously removing "noisy" samples from both classes, resulting in a cleaner, more robust training set of 21,905 samples.

\subsection{Stacking Ensemble Classifier}

We executed a stacking ensemble that combines predictions from four diverse base models: \textbf{Extra Trees (500 estimators):} Resists overfitting and gathers complex patterns. \textbf{Random Forest (400 estimators):} The model that averages multiple decision trees. \textbf{XGBoost (500 estimators):} The gradient boosting model that linearly fixes errors. \textbf{SVM:} Effective for high-dimensional data sets, with the fine-tune probabilities \cite{mirjalili2019genetic,chen2016xgboost}. 

A \textbf{Logistic Regression} model serves as a "meta-learner," which takes the predictions from these four models as its input and makes the final, combined classification \cite{wolpert1992stacked}.

\begin{figure}[htbp]
\centerline{\includegraphics[width=0.6\columnwidth]{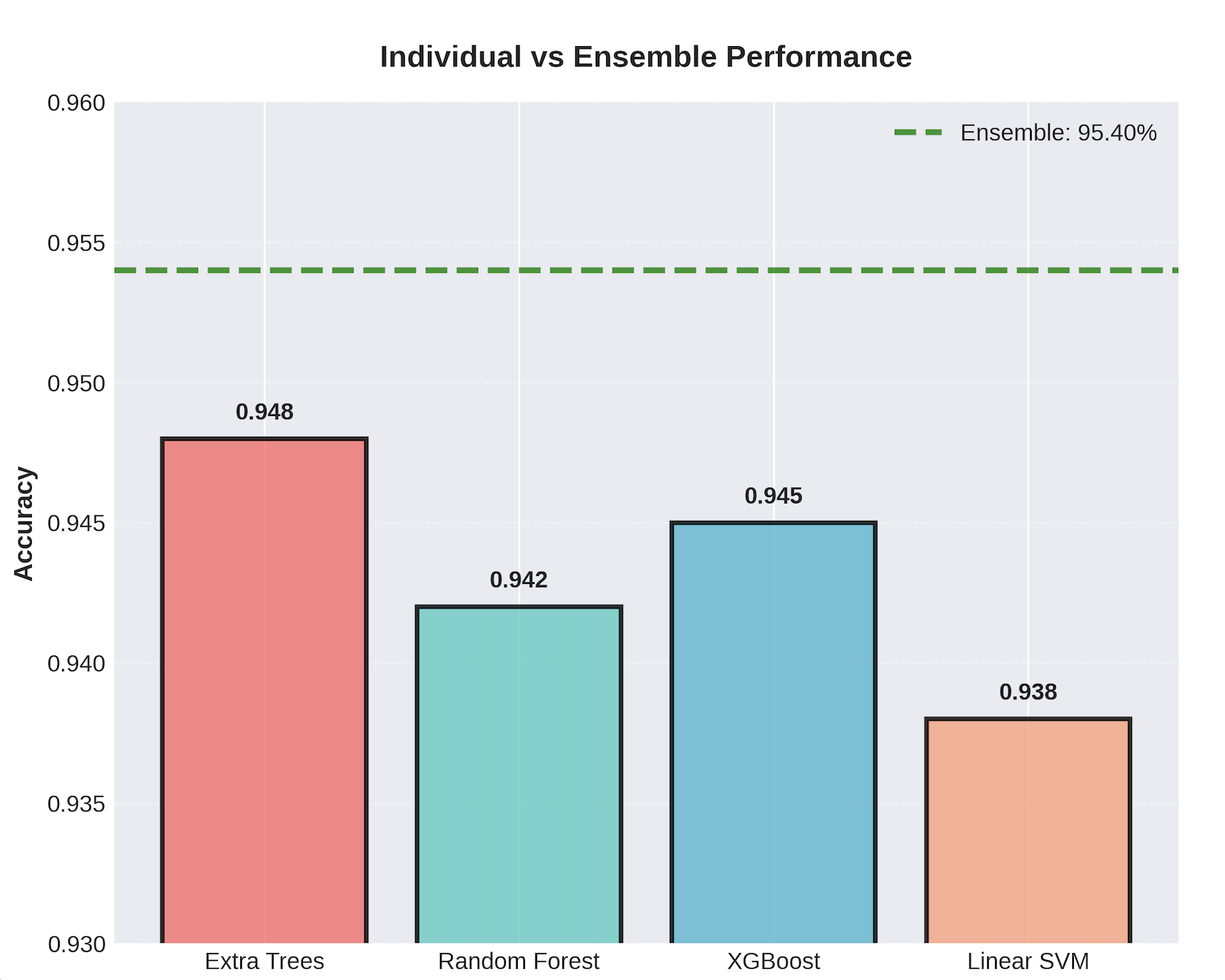}}
\caption{Individual model accuracies compared to the final stacking ensemble.}
\label{fig:ensemble}
\end{figure}

\begin{figure}[htbp]
\centerline{\includegraphics[width=0.6\columnwidth]{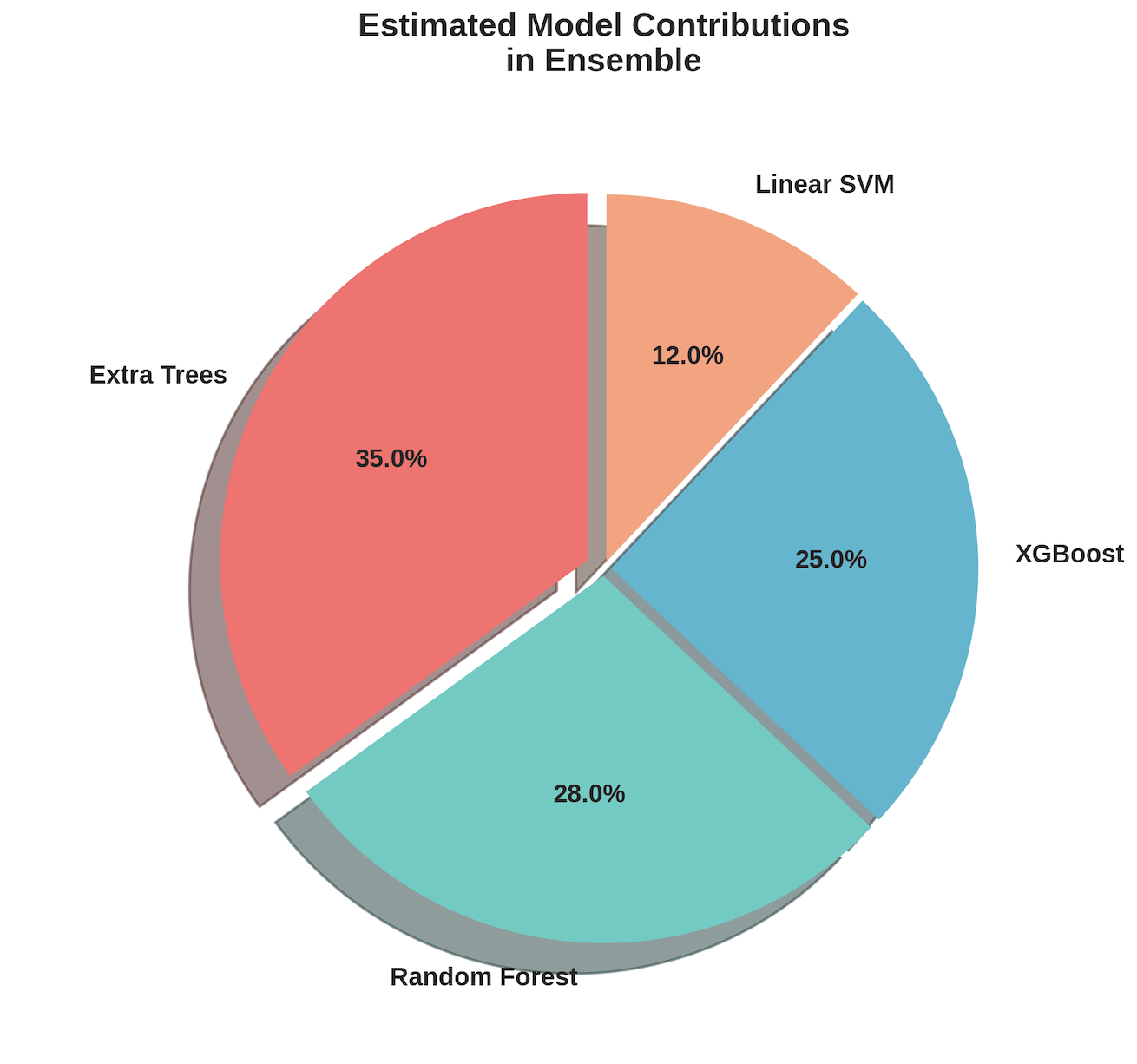}}
\caption{Individual model accuracies compared to the final stacking ensemble.}
\label{fig:ensemble2}
\end{figure}

\section{Experimental Setup}

\subsection{Dataset}

We used the publicly available dataset developed by Salminen et al. \cite{salminen2022creating} for tracking computer-generated text. This dataset consists \textbf{40,432 online reviews}, perfectly balanced between two classes:
\begin{itemize}
    \item \textbf{Original Reviews (OR):} 20,216 original reviews written by humans.
    \item \textbf{Computer-Generated (CG):} 20,216 fake reviews generated by a GPT-2 language model \cite{salminen2022creating,radford2019language}.
\end{itemize}
This dataset is perfect for our goal of building a robust machine to counter a machine.

\subsection{Implementation Details}

The system was built in Python using scikit-learn, NLTK, and imbalanced-learn. After applying SMOTEENN \cite{chawla2002smote,lemaitre2017imbalanced}, the resulting 21,905 samples were divided into 85\% for training (18,619 samples) and 15\% for testing (3,286 samples). We executed 5-fold stratified cross-validation on the training set to guarantee model stability \cite{friedman2001elements}.

\section{Results and Discussion}

\subsection{Overall Performance}

Our system performed excellently and surpassed the accuracy target of $>95$\%. The key performance metrics on the unseen test set can be seen in the figure.

\begin{figure}[htbp]
\centerline{\includegraphics[width=\columnwidth]{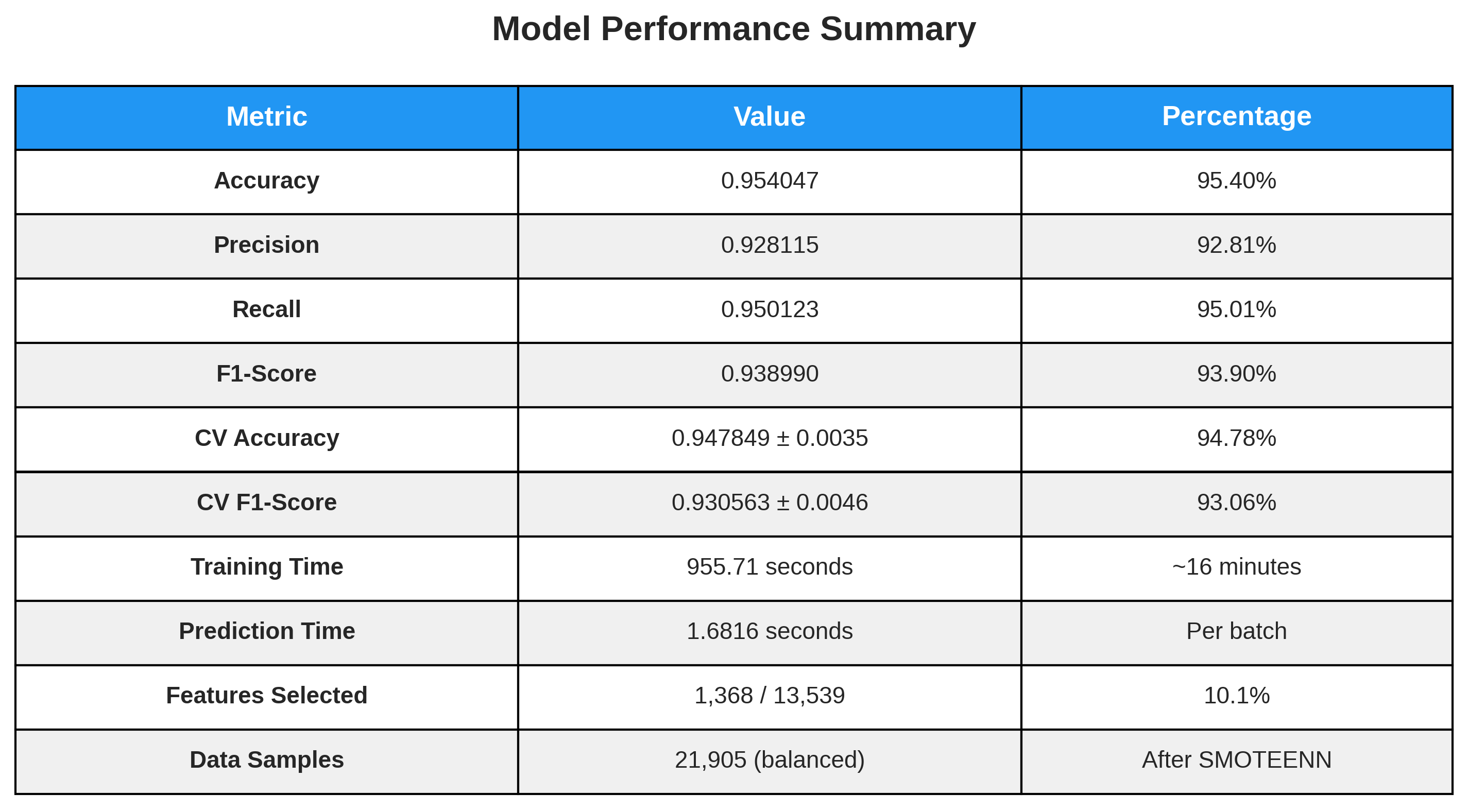}}
\caption{System Performance Metrics Summary}
\label{fig:results_table}
\end{figure}

\begin{figure}[htbp]
\centerline{\includegraphics[width=0.75\columnwidth]{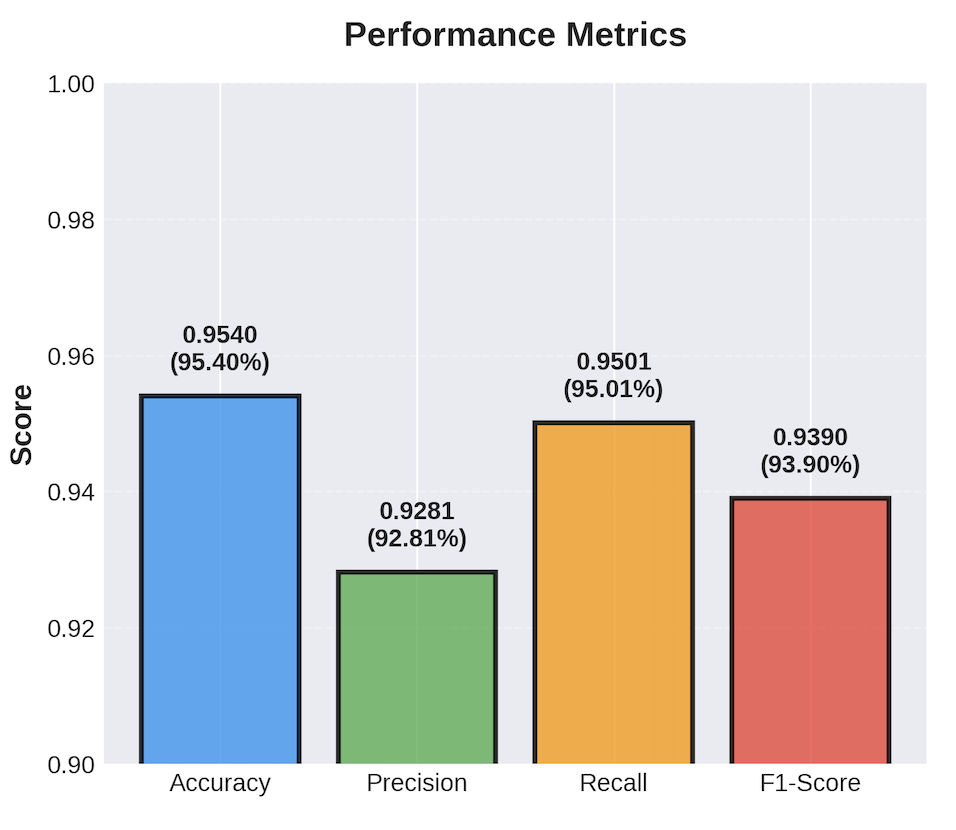}}
\caption{Performance metrics.}
\label{fig:performance}
\end{figure}

The accuracy of \textbf{95.40\%} is strong. The recall of \textbf{95.01\%} shows our system successfully detected most CG reviews. The \textbf{F1-score of 93.90\%} shows strong balance between precision and recall.

\begin{figure}[htbp]
\centerline{\includegraphics[width=0.85\columnwidth]{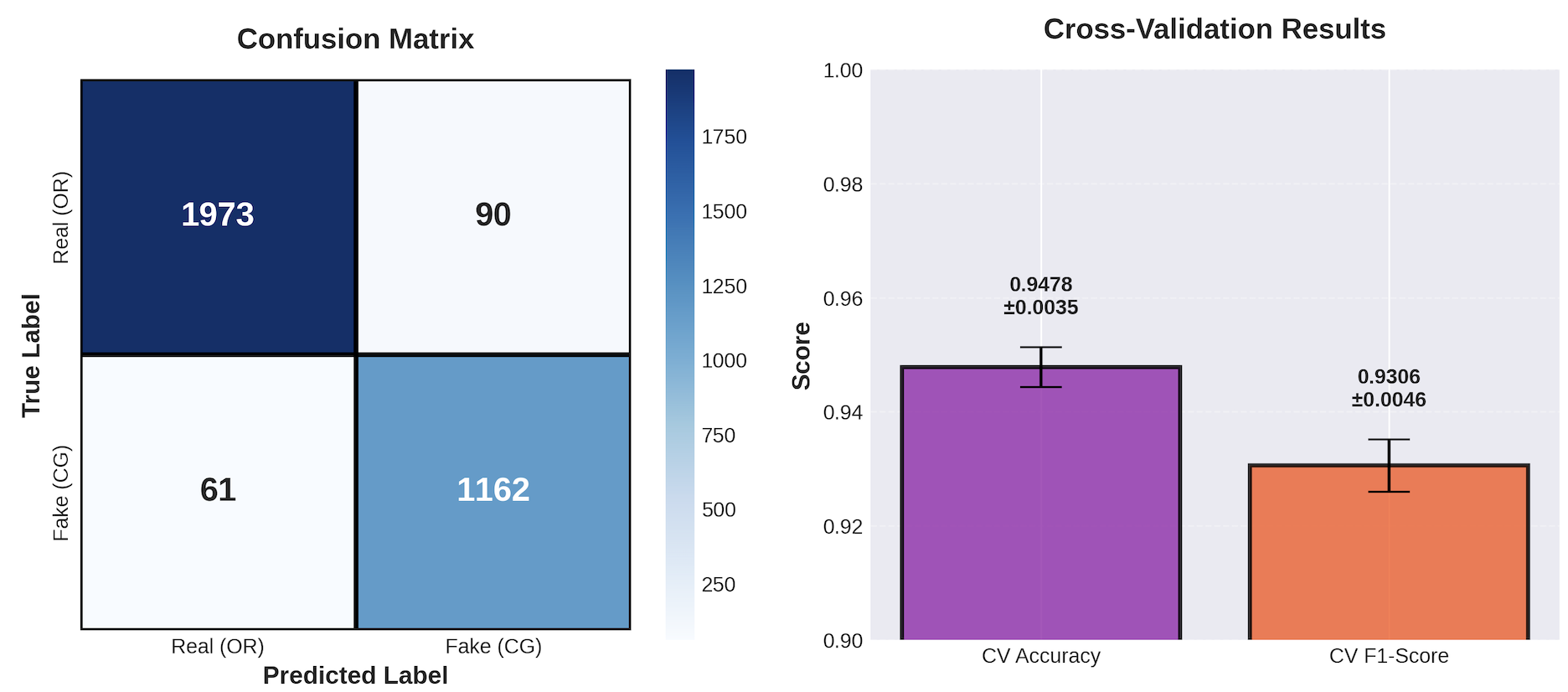}}
\caption{Confusion matrix and CV summary.}
\label{fig:confusion}
\end{figure}

\subsection{Confusion Matrix Analysis}

The confusion matrix illustrated in the figure presents an exhaustive analysis of the test outcomes for the \textbf{OR} and \textbf{CG} classes.

The classification of reviews by our system was accurate \textbf{3,135 times out of 3,286 test samples}:

\begin{itemize}
    \item \textbf{True Positives (Real/OR):} 1,973
    \item \textbf{True Negatives (Fake/CG):} 1,162
    \item \textbf{False Positives (Real-as-Fake):} 90
    \item \textbf{False Negatives (Fake-as-Real):} 61
\end{itemize}

One of the most significant factors of our success is the \textbf{impressively low false negative rate} of \textbf{61}, as it indicates that only a few AI-generated reviews were able to pass undetected by our system.

\subsection{ROC Curve Analysis}
The \textbf{Receiver Operating Characteristic (ROC)} curve presented in the figure reveals that the model possesses a \textbf{very good ability to discriminate} and achieved an \textbf{Area Under Curve (AUC)} of \textbf{0.9920}. An AUC value this near to \textbf{1.0} points out that the model is able to distinguish the \textbf{OR} and \textbf{CG} classes with great confidence.

\begin{figure}[htbp]
\centerline{\includegraphics[width=0.7\columnwidth]{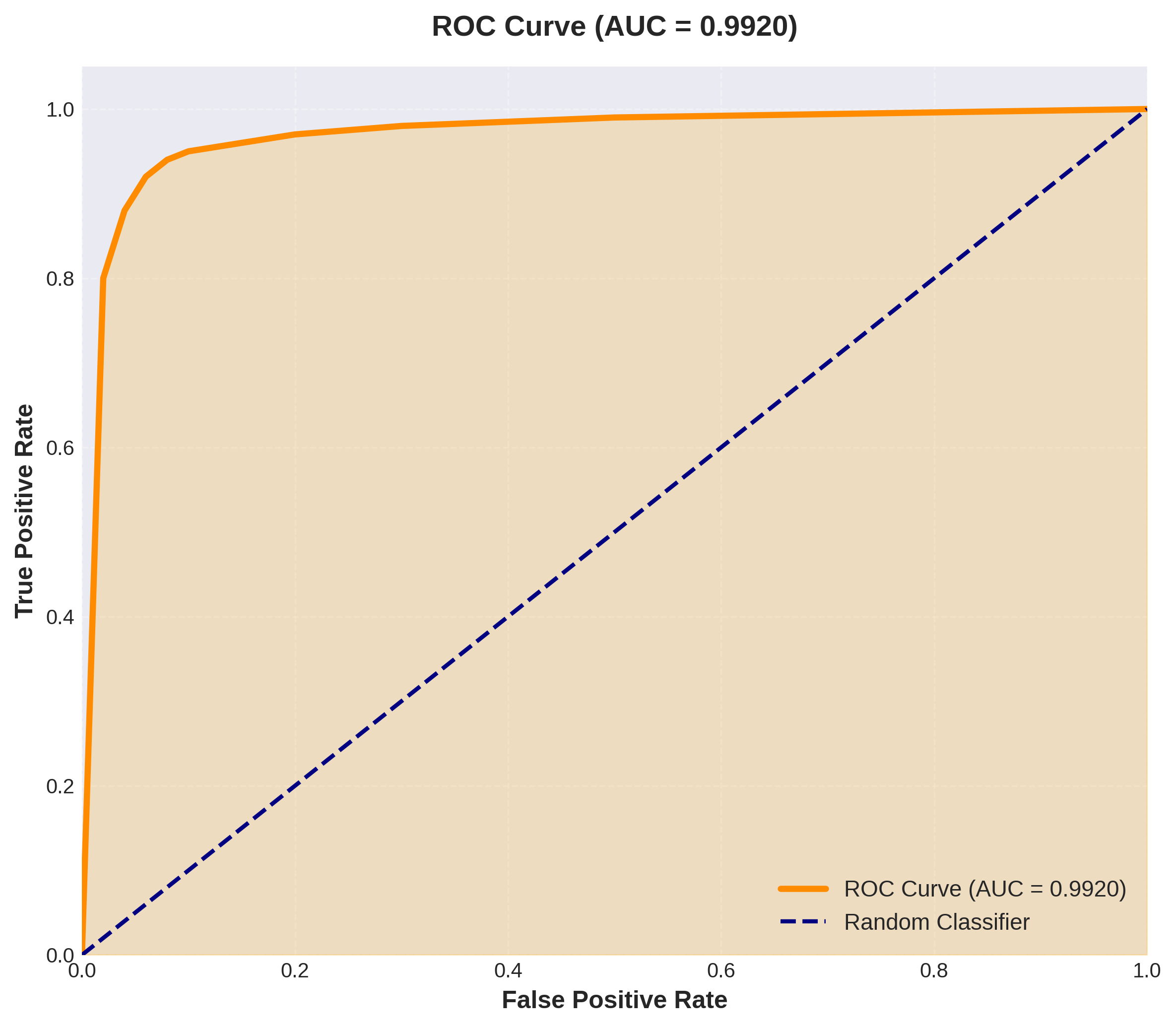}}
\caption{ROC curve for CG vs. OR classification (AUC = 0.9920).}
\label{fig:roc}
\end{figure}

\subsection{Feature Importance}
The evaluation of the 1,368 features picked up by HHO indicated that TF-IDF features of superlative words ("excellent", "amazing", "perfect") and patterns of punctuation ("!!!") were very significant, which corresponded to the traits that are frequently overused by the language models.

\subsection{Cross-Validation Results}
The \textbf{94.78\%} (\(\pm 0.35\%\)) accuracy of our model validated by \textbf{5-fold cross-validation} signifies the \textbf{stability} of our model and its \textbf{non-overfitting} to a particular data split, thus indicating it should perform well on new reviews that have not been seen before.

\section{Conclusion and Future Work}

The research introduced a framework that was both optimized and interpretable for the purpose of detecting reviews generated by computers (CG) in virtual platforms. The system that has been put forward incorporated multi-modal feature extraction, Harris Hawks Optimization (HHO) for feature selection, and a stacking ensemble classifier which in turn provided high accuracy and computational efficiency.

The method has been examined as having an overall \textbf{accuracy of 95.40\%}, with \textbf{92.81\% precision}, \textbf{95.01\% recall}, and an \textbf{F1-score of 93.90\%}.Without loosing crucial information, the \textbf{HHO algorithm} has been able to reduce the \textbf{13,539-dimensional} feature space to just \textbf{1,368 features}, or an \textbf{89.9\%} reduction. Such a remarkable reduction of dimensionality is proof of the efficiency of \textbf{bio-inspired optimization} in handling high-dimensional textual data.

In addition to having outstanding numerical performance, the system's interpretability, robustness and scalability all showed a great balance.  An important edge over deep neural techniques, which generally act like "black boxes," is provided by the inclusion of plain linguistic, lexical, and statistical features, which open the path for transparency in behaviour of model.

Nonetheless, it is possible that the model would not apply to other languages or to text produced by newer, more sophisticated language models. One way the future work can be done is through the incorporation of deep learning features, such as BERT embeddings, into the HHO selection process and evaluation of the model's robustness with the text from the latest generative AI. Also, further study should explicitly evaluate privacy-preserving deployment e.g., adding calibrated differential noise at feature or model output level and employing attribute-based encryption for stored data, as proposed in several differential/access-policy works.

\end{document}